\documentclass[journal,twoside,web]{ieeecolor}
\usepackage{jsen}
\usepackage{cite}
\usepackage{amsmath,amssymb,amsfonts}
\usepackage{algorithmic}
\usepackage{graphicx}
\usepackage{textcomp}
\usepackage{wrapfig}
\usepackage[acronym]{glossaries}
\usepackage{xcolor}

\makeglossaries
\newacronym{vbts}{VBTS}{Vision-Based Tactile Sensor}
\newacronym{mbt}{MBT}{Marker-Based Transduction}
\newacronym{smb}{SMB}{Simple Marker-Based}
\newacronym{mmb}{MMB}{Morphological Marker-Based}
\newacronym{ibt}{IBT}{Intensity-Based Transduction}
\newacronym{rlb}{RLB}{Reflective Layer-Based}
\newacronym{tlb}{TLB}{Transparent Layer-Based}

\def\BibTeX{{\rm B\kern-.05em{\sc i\kern-.025em b}\kern-.08em
    T\kern-.1667em\lower.7ex\hbox{E}\kern-.125emX}}
\definecolor{abstractbg}{rgb}{0.89804,0.94510,0.83137}
\begin{document}
\title{Classification of Vision-Based Tactile Sensors:\\ A Review}
\author{Haoran Li, Yijiong Lin, Chenghua Lu, Max Yang, Efi Psomopoulou, Nathan F. Lepora 
\thanks{CL, HL and YL were supported by the the China Scholarship Council and Bristol joint scholarship. EP and NL were supported by the Horizon Europe research and innovation program under grant agreement No. 101120823 (MANiBOT) and the Royal Society International Collaboration Awards (South Korea). NL was also supported by an award from ARIA on `Democratising Hardware And Control For Robot Dexterity'. {\em (Corresponding author: Nathan F. Lepora)}}
\thanks{HL is with School of Robotics, Xi'an Jiaotong-Liverpool University, China, and was with the School of Engineering Mathematics and Technology, and Bristol Robotics Laboratory, University of Bristol, Bristol, U.K. (Email: haoran.li@xjtlu.edu.cn).}
\thanks{YL, CL, MY, EP, and NL are with the School of Engineering Mathematics and Technology, and Bristol Robotics Laboratory, University of Bristol, Bristol, U.K. (Email: \{yijiong.lin, chenghua.lu, max.yang, efi.psomopoulou, n.lepora\}@bristol.ac.uk).}
}

\IEEEtitleabstractindextext{%
\fcolorbox{abstractbg}{abstractbg}{%
\begin{minipage}{\textwidth}%
\begin{wrapfigure}[21]{r}{3in}%
\includegraphics[width=2.8in]{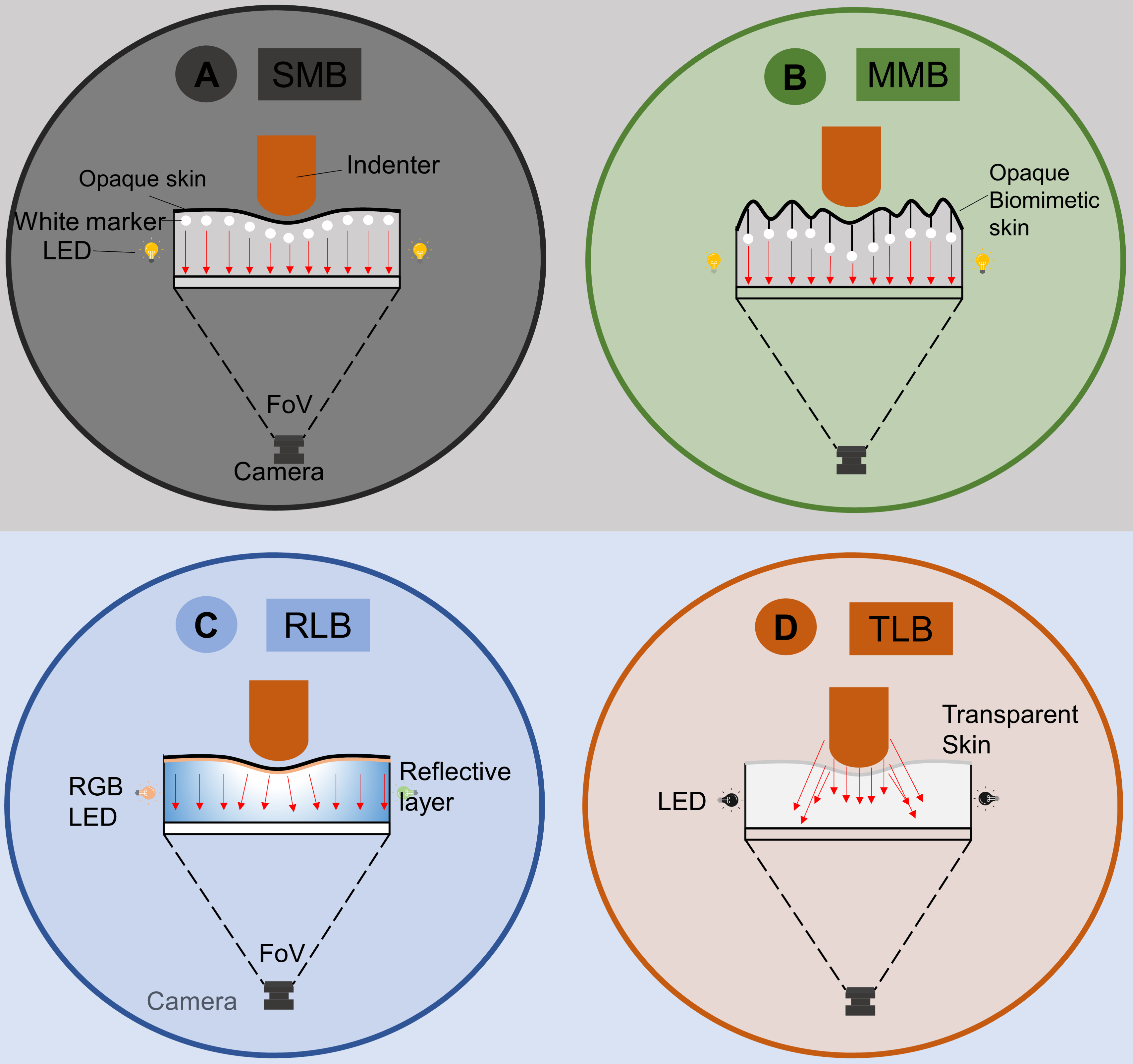}%
\end{wrapfigure}%
\begin{abstract}
Vision-based tactile sensors (VBTS) have gained widespread application in robotic hands, grippers and prosthetics due to their high spatial resolution, low manufacturing costs, and ease of customization. While VBTSs have common design features, such as a camera module, they can differ in a rich diversity of sensing principles, material compositions, multimodal approaches, and data interpretation methods. Here, we propose a novel classification of VBTS that categorizes the technology into two primary sensing principles based on the underlying transduction of contact into a tactile image: the Marker-Based Transduction Principle and the Intensity-Based Transduction Principle. Marker-Based Transduction interprets tactile information by detecting marker displacement and changes in marker density. In contrast, Intensity-Based Transduction maps external disturbances with variations in pixel values. Depending on the design of the contact module, Marker-Based Transduction can be further divided into two subtypes: Simple Marker-Based (SMB) and Morphological Marker-Based (MMB) mechanisms. Similarly, the Intensity-Based Transduction Principle encompasses the Reflective Layer-based  (RLB) and Transparent Layer-Based (TLB) mechanisms. This paper provides a comparative study of the hardware characteristics of these four types of sensors including various combination types, and discusses the commonly used methods for interpreting tactile information. This~comparison reveals some current challenges faced by VBTS technology and directions for future research.
\end{abstract}

\end{minipage}}}

\maketitle

\section{Introduction}
\label{sec:introduction}
In robotic systems, tactile sensing is fundamental for enabling robots to interact with their environment through physical contact. \textcolor{black}{By delivering real-time tactile feedback, such as object stiffness, local force, slip and contact position feedback, this capability empowers robotic systems to achieve precise object manipulation while preventing damage~\cite{howe1993tactile,yousef2011tactile,luo2017robotic,chen2018tactile}. Such functionality proves critical for applications ranging from VR/AR teleoperation systems to assistive prosthetic devices designed for individuals with physical disabilities, where accurate environmental interaction and haptic awareness are operationally essential.} 

\textcolor{black}{Traditional electronic technologies such as piezoelectric and piezoresistive sensor arrays have been considered promising due to their high temporal resolution and thin profiles. However, their utility is constrained by their limited spatial resolution ~\cite{nicholls1989survey, lucarotti2013synthetic}, which impedes their ability to capture detailed contact information, including fine shape features, object textures, and multidimensional contact forces.} Furthermore, the process of customizing these sensors to new designs can require significant effort to redistribute the unit arrays and repackage the associated circuitry~\cite{zhu2022recent, shu2014new, yang2013gauge, zhou2021three}, which presents a barrier to integrating and adopting them in robotic manipulation systems~\cite{dahiya2013directions}. Consequently, their application to state-of-the-art dexterous robotic tasks has been limited.

\glspl{vbts} have become popular in robotic hands and grippers, mainly due to their high spatial resolution, low production costs, and ease of customization~\cite{gao2022tactile}. Compared to other tactile sensing technologies, \glspl{vbts} are characterized by a relatively simple design, typically comprising three essential components: a soft-skinned contact module, an illumination module, and a camera module~\cite{zhang2022hardware}. The contact module is integral to the transduction of contact into a visible pattern that can be imaged. Typically, it can be rapidly developed and fabricated using accessible techniques like silicone molding~\cite{lin20239dtact} or multi-material 3D printing~\cite{lepora2021soft}, which has enabled many different designs to emerge. This ease and flexibility in design not only shortens production times, but also enhances customization potential, enabling \glspl{vbts} to be applied to many diverse applications. In fields like surgery and healthcare, VBTS-equipped robots can perform rapid object manipulation and, when combined with haptic feedback, can enable precise, tactile-assisted procedures~\cite{tiwana2012review}. This tactile feedback can improve both the safety and dexterity of robotic systems in healthcare and service industries, facilitating more delicate and secure interactions with tools and objects~\cite{konstantinova2014implementation,girao2013tactile}.


The output of a \gls{vbts} is a tactile image. Typically, this is captured by a camera module in conjunction with an illumination module, which captures images of the internal sensing region when subjected to external stimuli. These high-resolution images can then be processed to extract rich tactile information, for which a variety of interpretation algorithms have been developed~\cite{yamaguchi2019recent, li2023marker}. Researchers leverage advanced machine learning techniques\textcolor{black}{~\cite{lloyd2023pose,fan2024vitactip, subad2021soft}} and physical modeling~\cite{li2024biotactip, yuan2017gelsight} to extract diverse tactile information from raw marker displacement fields~\cite{li2023marker}. By analyzing these displacement patterns, they can decode contact characteristics such as force distribution, pressure, and texture. This information allows robots to sense and respond to complex shapes and interactions with high accuracy~\cite{li2020review}. Such advantages make \glspl{vbts} highly efficient in improving the tactile perception of robotic systems while also ensuring cost-effectiveness and customizability. Consequently, \glspl{vbts} have emerged as highly versatile and cost-effective solution for a broad range of robotic tasks, from precision manipulation~\cite{bauza2024simple} to advanced object recognition~\cite{liu2016visual, gandarias2019cnn, liu2017recent}. 

\textcolor{black}{This paper is organized as follows: Section II discusses review articles related to \glspl{vbts}. Section III introduces the definition and main components of \glspl{vbts} and presents the primary sensing principles of \glspl{vbts} and its corresponding subtypes. Section IV describes \glspl{vbts} with multiple perception mechanisms. Section V discusses data processing methods for \glspl{vbts}. Section VI discusses the current challenges for \glspl{vbts}. Finally, Section VII concludes this paper with a summary of the main findings.}

\section{Related Works}
In recent years, a succession of reviews have introduced advances in \glspl{vbts}. Li et al. (2024)~\cite{li2024vision} offered a thorough review of visuotactile sensors from a signal processing perspective, emphasizing the critical role of techniques such as contact area segmentation, force perception, and 3D reconstruction. Li et al. (2023)~\cite{li2023marker} reviewed marker displacement methods (MDM) in \glspl{vbts}, categorizing them into 2D, 2.5D, and 3D approaches, providing insights into enhancing the interpretation techniques for MDM-based tactile sensors and the use of multiple cameras in a \glspl{vbts}. Zhang et al. (2022)~\cite{zhang2022hardware} focused on \gls{vbts} hardware technologies related to contact, illumination, and camera modules, to highlight key challenges in durability, sensitivity, and multimodal integration to improve future \gls{vbts} designs. Lepora (2021)~\cite{lepora2021soft} focused on soft biomimetic optical tactile sensing, exemplified by the TacTip's biomimicry of the structure and function of human glabrous skin, and the key role of tactile shear sensing for future robot dexterity. Abad and Ranasinghe (2020)~\cite{abad2020visuotactile} conducted an in-depth review of the GelSight sensor, emphasizing its applications in robotics, haptics, and computer vision, and showcasing its ability for real-time image analysis and 3D reconstruction. Ward-Cherrier et al. (2018)~\cite{ward2018tactip} described a family of 3d-printed biomimetic tactile sensors based on the TacTip design that use biomimetic markers to enhance the sensitivity to contact. Yuan et al. (2017)~\cite{yuan2017gelsight} covered the design elements of the GelSight, encompassing its optical system, design structure, algorithms to measure force and slip, and hardware improvements made to better integrate with robotic platforms. 

\textcolor{black}{Within these reviews were multiple attempts to categorize \glspl{vbts} using various classification frameworks. Shimonomura (2019)~\cite{shimonomura2019tactile} classified camera-based tactile image sensors into three types based on differences in the contact module's hardware: light-conductive plate-based sensors, marker displacement-based sensors, and reflective membrane-based sensors. Similarly, Shah et al. (2021) \cite{shah2021design} categorized \glspl{vbts} into three design paradigms: waveguide-type designs, marker displacement-based designs, and reflective membrane designs. Although both of these classifications distinguish between marker-based and reflective-based approaches, they overlook designs that combine these principles. Also, as argued here, marker-based designs are usefully distinguished further, while reflective-based types represent a subcategory of \glspl{vbts}. In addition, Shah et al. highlighted the significance of waveguide-type designs, such as optical fibers, but this consideration is distinct from how contact is transduced in a tactile image due to the mechanics of the skin~\cite{shah2021design}.}

\textcolor{black}{Upon reviewing the literature and considering the rapid progress and proliferation of new \glspl{vbts},  we observed a notable absence of a unified framework to classify the different types of \gls{vbts}. The basic categorization into marker-based and reflection-based VBTS, which was adequate five years ago, no longer captures the complexity of contemporary designs. This lack of a unified characterization of the technology has led to challenges in systematically categorizing and evaluating the expanding array of \glspl{vbts} technologies. As novel designs increasingly integrate multiple sensing mechanisms, material innovations, multimodalities, and interpretation methods, the categorization of marker-based sensors versus reflection-based sensors has become insufficient to capture and guide progress in \glspl{vbts} technology.}

\begin{figure*}
    \centering
    \includegraphics[width=1\linewidth]{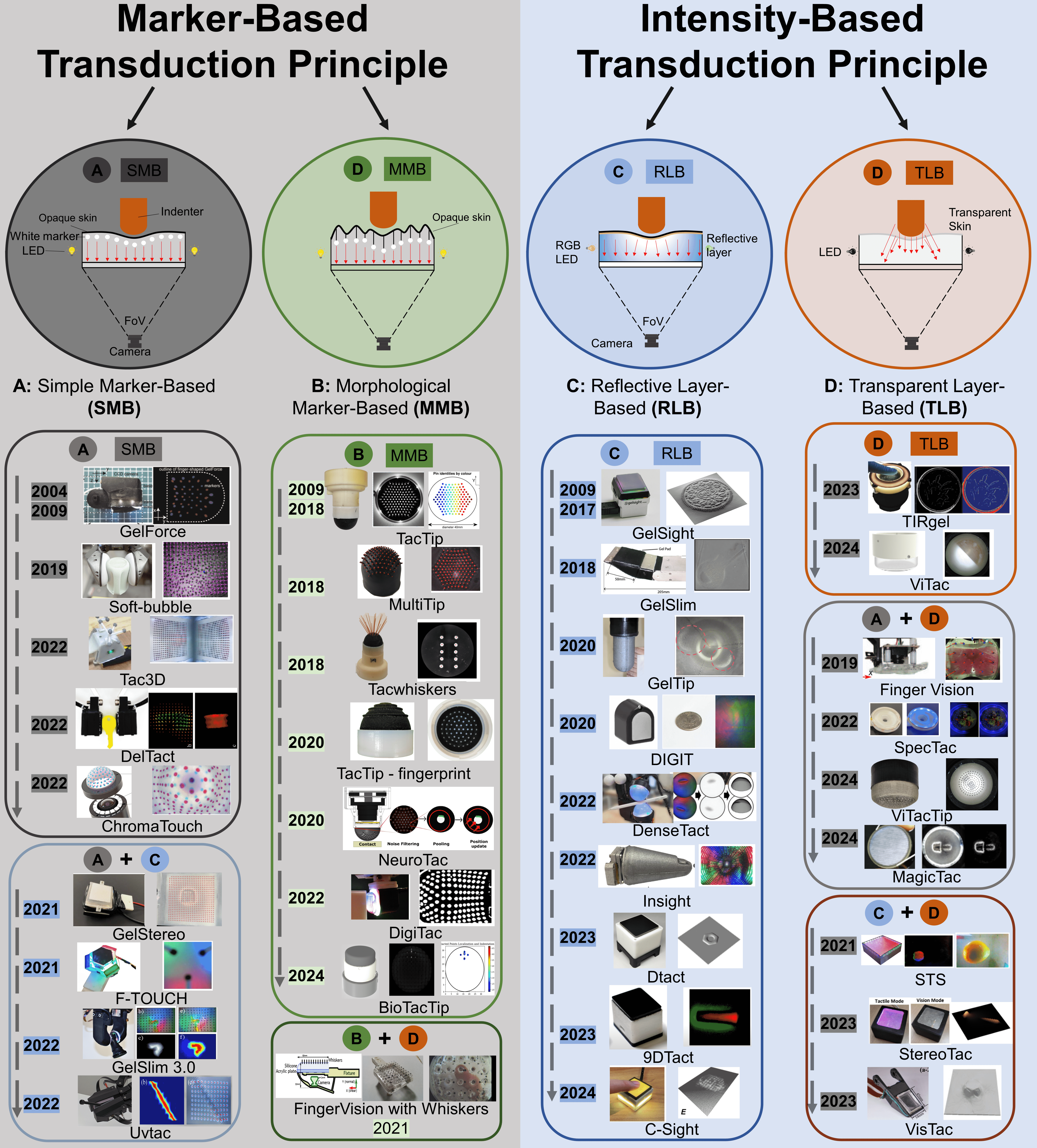}
    \caption{Typical \gls{smb} sensors include ChromaTouch~\cite{scharff2022rapid}, GelForce ~\cite{kamiyama2004gelforce, sato2009finger}, Tac3D~\cite{zhang2022tac3d},  DelTact~\cite{zhang2022deltact} and Soft-bubble~\cite{alspach2019soft, kuppuswamy2020soft}. \gls{mmb} sensors include TacTip~\cite{chorley2009development, lepora2021soft, ward2018tactip}, MultiTip~\cite{soter2018multitip}, DigiTac~\cite{lepora2022digitac}, TacWhiskers~\cite{lepora2018tacwhiskers}, TacTip with fingerprint~\cite{james2020biomimetic}, BioTacTip~\cite{li2024biotactip} and NeuroTac~\cite{ward2020neurotac}. \gls{rlb} Sensors include GelSight~\cite{yuan2017gelsight,johnson2009retrographic}, DIGIT~\cite{lambeta2020digit}, 9DTact~\cite{lin20239dtact}, C-Sight~\cite{fan2024design}, GelSlim~\cite{donlon2018gelslim}, GelTip~\cite{gomes2020geltip},DenseTact~\cite{do2022densetact}, Insight~\cite{sun2022soft}, Dtact~\cite{lin2023dtact}. \gls{tlb} Sensors include TIRgel~\cite{zhang2023tirgel} and ViTac~\cite{fan2024vitactip}. \textcolor{black}{\gls{smb}+\gls{rlb} (A+C) sensors incorporate simple markers within its reflective layer}. Typical \gls{smb}+\gls{rlb} (A+C) sensors include GelStereo~\cite{cui2021hand}, F-Touch~\cite{li2020f}, GelSlim 3.0~\cite{taylor2022gelslim} and UVtac~\cite{kim2022uvtac}. \textcolor{black}{\gls{mmb}+\gls{tlb} (B+D) sensors feature a morphologically patterned tactile structure embedded within its transparent skin, distinguishing it from simple marker-based designs.} \gls{mmb}+\gls{tlb} (B+D) sensor includes FingerVision with Whiskers~\cite{yamaguchi2021fingervision}.
    \textcolor{black}{\gls{smb}+\gls{tlb} (A+D) sensors lie in the incorporation of markers within its transparent skin.} Representative \gls{smb}+\gls{tlb} (A+D) sensors include ViTacTip~\cite{fan2024vitactip}, MagicTac~\cite{fan2024magictac}, Finger Vision~\cite{Yamaguchi19IJHR_FingerVision} and SpecTac~\cite{wang2022spectac}.  \textcolor{black}{\gls{rlb}+\gls{tlb} (C+D) sensors feature a transparent outer layer integrated with a reflective layer.} \gls{rlb}+\gls{tlb} (C+D) sensors include STS~\cite{hogan2021seeing}, StereoTac~\cite{roberge2023stereotac} and VisTac~\cite{athar2023vistac}.}
    \label{fig:optical_sensors_category}
\end{figure*}

To address this gap, we propose an innovative classification based on the sensory transduction principles. We first divide \glspl{vbts} into two primary transduction principles: \gls{mbt} and \gls{ibt}, depending on the detection of the marker displacement or density changes and pixel intensity variations to interpret tactile information \textcolor{black}{separately}. These transduction principles are then subdivided into two further mechanisms. The \gls{mbt} principle separates into \gls{smb} and \gls{mmb}. SMB only uses discrete markers to convey deformation, whereas MMB integrates markers with specialized geometries to enhance tactile feedback. The \gls{ibt} principle is divided into \gls{rlb} and \gls{tlb}. RLB employs reflective layers to highlight variations in light when it interacts with the surface, whereas TLB uses transparent skin to enable multiple modalities that can capture both high-resolution contact information and proximal details of nearby objects, such as their texture. Furthermore, we will see many examples where these transduction principles and subclasses are combined within a single \gls{vbts}, but nevertheless the classification allows us to characterize the component transduction methods. In general, our classification provides researchers in the field with a clear framework, helping them quickly understand the evolution and technical branches of \glspl{vbts} and we intend it will inspire innovation of new technologies.





\section{Basic classification scheme for \acrlong{vbts}s}\label{sec:VBTS_hardware}
\subsection{Defining Vision-Based Tactile Sensing}\color{black}  

\begin{figure*}
    \centering
    \includegraphics[width=0.9\linewidth]{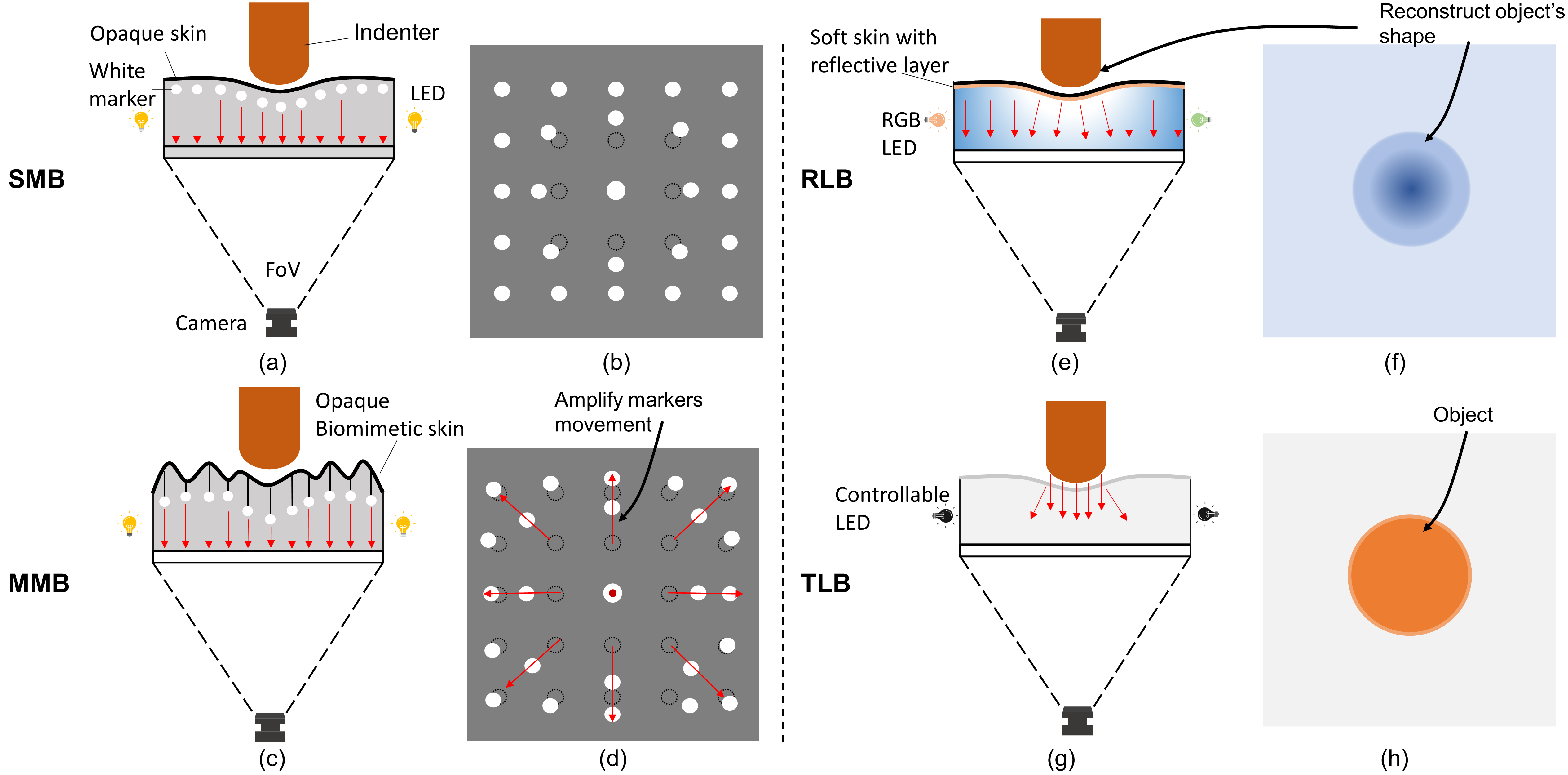}
    \caption{Schematic camera views of SMB, MMB, RLB, and TLB sensors. (b) Shift in marker position upon indentation of SMB, (d) Amplified marker movement, (f) reconstructed object shape, (h) visual and tactile information.}
    \label{fig:camera_view}
\end{figure*}
Here we consider an idealized \gls{vbts} to be composed of four modules (see Fig.~\ref{fig:optical_sensors_category} and Table~\ref{tab:main componsents}): a contact module, an illumination module, a perception module, and a base module. The contact module is designed to interact directly with the external environment and is typically crafted from soft, pliable materials such as silicone or rubber. This flexibility allows the module to conform to the shapes and textures of various objects, effectively translating physical contacts into patterns that the VBTS can interpret. The soft materials in the contact module help to enhance sensitivity and reduce damage to objects, making VBTS particularly suitable for delicate handling in unstructured settings.

The illumination module provides the necessary lighting to ensure high-quality image capture. This module often includes white or RGB LEDs, or in some cases, UV lights, to create consistent illumination across the contact surface. Reliable lighting is critical to producing clear, high-contrast images of contact patterns. RGB lighting, for instance, can enhance image differentiation and provide richer data on the contact, while UV lighting might highlight specific materials or features that are not visible in standard lighting conditions. 

At the heart of the VBTS is the perception module, which houses the cameras and any required structural supports. This module captures high-resolution images of contact events, allowing for detailed data acquisition. In advanced designs, this module may also incorporate an on-board tactile model or algorithms that process raw images in real time. Such models interpret the contact images, translating visual information into useful tactile data like contact force, pressure distribution, or texture, which enhances the sensor's ability to make accurate assessments on the spot. Finally, an optional base module can function to support the tactile sensor and interconnect the other modules, ensuring the system operates effectively as a whole.

\subsection{\acrlong{mbt}}
\label{sec:mbt}

\subsubsection{\acrlong{smb}}
The \gls{smb} transduction mechanism interprets the sensor-surface contact deformation by observing the displacement of markers under an external disturbance~\cite{li2023marker} (see Fig.~\ref{fig:camera_view}). The key characteristic of this type of \gls{vbts} is the usage of markers generally made up of a structured array of floating points to transmit tactile information. Of the four subdivided \gls{vbts} types, \gls{smb} is the most straightforward and simplest to produce, and has many variations (Fig.~\ref{fig:optical_sensors_category}-A).



The original GelForce sensor, introduced by~Kamiyama et al.~\cite{kamiyama2005vision} in 2005, has since undergone several modifications, particularly in its size and shape~\cite{sato2009finger}. GelForce is composed of two layers of markers, which typically include red and blue markers to track deformation. A single camera captures the relative movement of these marker layers, enabling the system to interpret tactile information, such as contact forces from the relative marker displacements. More recently, the Soft-Bubble sensor~\cite{alspach2019soft, kuppuswamy2020soft} incorporates a layer of randomly-distributed high-density internal markers on the inner surface of the bubble to track dense optical flow patterns that indicate membrane displacement. Tac3D~\cite{zhang2022tac3d} employs stereo cameras to capture tactile images, enhancing its depth prediction capabilities and helping reconstruct object shapes. DelTact~\cite{zhang2022deltact} employs a dense color pattern to capture tactile information across its sensing surface to track pixel-level changes within this color pattern, providing the distribution of contact forces. \textcolor{black}{ChromaTouch~\cite{scharff2022rapid} consists of two distinctly-colored red and blue marker layers that shift displacement under external pressures, blending the hue of the markers when they overlap. This novel mechanism gives more information about the relative positions of these markers in addition to the changes in marker positions on the tactile image. }


\subsubsection{\acrlong{mmb}}
\gls{mmb} transduction mechanisms enhance the \gls{smb} mechanism by mechanically increasing sensitivity or adding other sensing capabilities, such as biomimetic designs to the epidermis or within the contact module~\cite{yi2018biomimetic} (see Fig.~\ref{fig:camera_view}). Such designs can mimic natural tactile structures, improving the sensor's ability to detect finer details of contact forces, textures, or environmental conditions, enabling MMB sensors to be more adaptable to advanced applications in robotics (Fig.~\ref{fig:optical_sensors_category}-B).



\gls{mmb} sensors are usually equipped with biomimetic structures such as the TacTip series that originated in Bristol Robotics Laboratory in 2009~\cite{chorley2009development,ward2018tactip}, which features elongated pin tips with white markers at the ends attached to a flexible black surface. This design employs a micro-leverage effect to amplify deformations of the skin, thereby enhancing the sensor's sensitivity. Subsequently, several variations inspired by human sensory features have been developed. The TacTip-Fingerprint~\cite{james2020biomimetic}, for example, integrates multilayered grooves on its surface to mimic human fingerprints, improving its ability to detect shear forces. The MultiTip sensor~\cite{soter2018multitip}, on the other hand, uses thermochromic materials on its surface, enabling it to sense temperature changes. TacWhiskers features a rigid yet flexible tip, resembling mouse whiskers, which connects to internal markers, allowing the sensor to capture a broader range of environmental information. More recently, DigiTac~\cite{lepora2022digitac} modifies the shape of the TacTip to be compatible with the DIGIT base, giving a smaller form factor that enables applications that require integration into fingertips. The BioTacTip sensor improves functionality through the use of markers and cover tips~\cite{li2024biotactip}: in a non-contact state, the markers are invisible, then under external force in the contact area, the white tip emerges to locate the contact zone. The appearance of the white tip not only aids in positioning, but its reflected light intensity is directly proportional to the extent of tip emergence, which linearly correlates with contact force. This kind of sensor illustrates that MMB-based sensors, by designing and combining markers or tips, can transform the contact information through the morphology of the sensing surface while maintaining the other benefits of SMB sensors, such as simplicity and ease of fabrication. 

\begin{table*}
\centering
\caption{\textcolor{black}{Main components for different types of \glspl{vbts}}}
\label{tab:main componsents}
\begin{tabular}{c}
\includegraphics[width=1\textwidth]{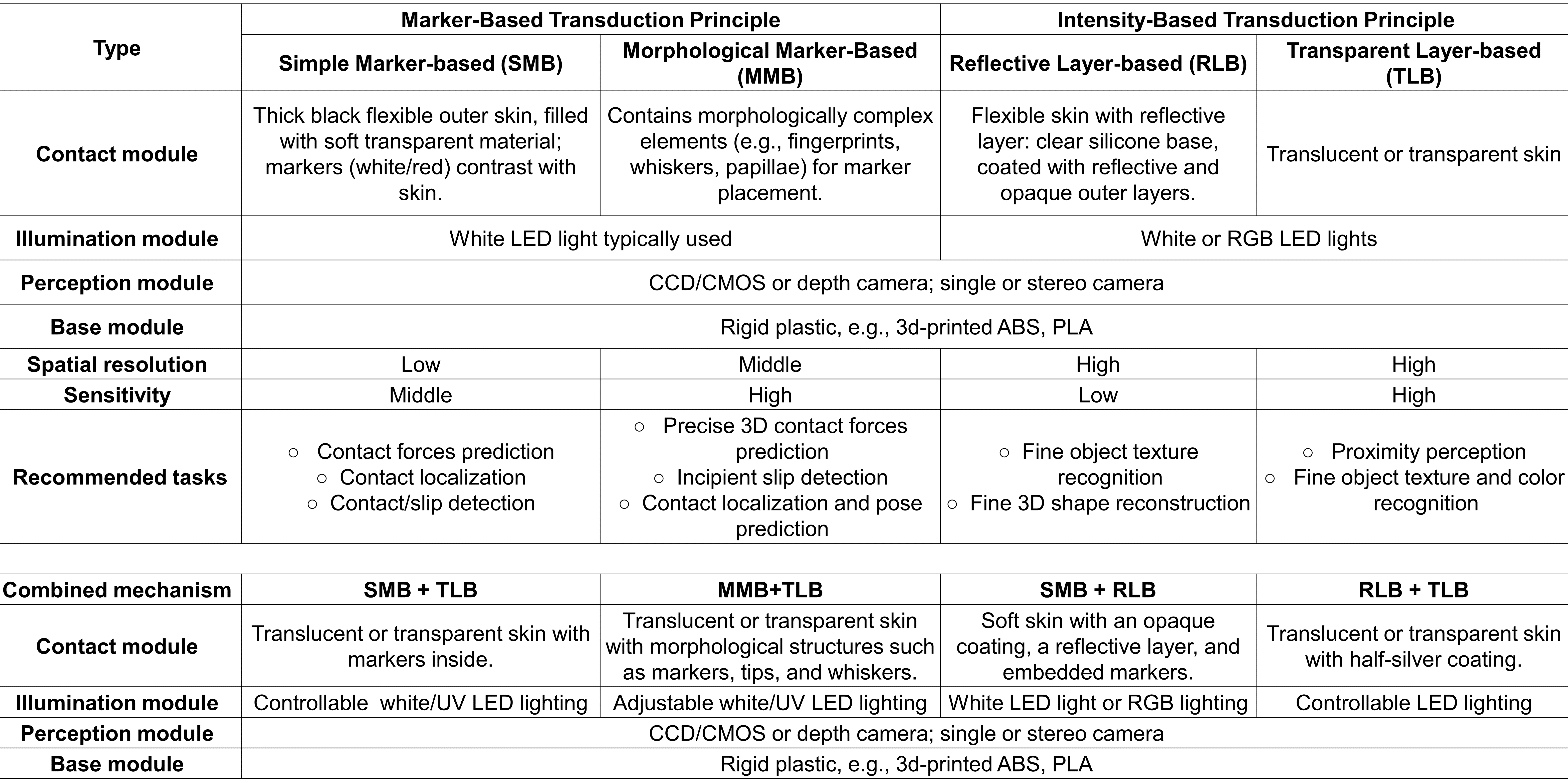} \\ 
\end{tabular}
\end{table*}


\subsection{\acrlong{ibt}}\color{black} 
\label{sec:intensity based sensing}
\subsubsection{\acrlong{rlb}}
The \acrlong{rlb} sensors transmit contact information by analyzing changes in light intensity resulting from contact (see Fig.~\ref{fig:camera_view}). To capture these changes in light intensity, RLB sensors are typically designed with multiple layers, including an outer layer and a reflective layer. The outer layer is usually made of opaque material to prevent external environmental light from entering the sensor, thereby ensuring the quality of the tactile imaging. The reflective layer is often coated with metallic paint to effectively reflect internal RGB light (see Fig.~\ref{fig:optical_sensors_category}-C). 


Among the most notable RLB sensors is the GelSight, originally developed within MIT in 2009~\cite{yuan2017gelsight,johnson2009retrographic}. GelSight's surface is covered with an opaque reflective layer and is illuminated using integrated RGB LEDs. It utilized a stereo photometric algorithm to generate a depth map, allowing for reconstruction of the shape and surface texture of the object it contacts. Building on the GelSight, the  GelSlim~\cite{donlon2018gelslim} incorporates a redesigned internal lighting pathway by integrating light guides and specular reflection arrangements. This innovative setup allows for a more compact and slim integration, achieving efficient illumination within a reduced form factor. GelTip~\cite{gomes2020geltip} adopts the same sensing principle as GelSight~\cite{yuan2017gelsight,johnson2009retrographic} but innovatively features a 3D cylindrical sensing surface. By incorporating a specially designed projective function, GelTip can detect contact from multiple directions, significantly enhancing its tactile perception capabilities. Another variant, the DIGIT has reduced the sensor size and eased fabrication while preserving its tactile capabilities~\cite{lambeta2020digit}. Meanwhile, DenseTact~\cite{do2022densetact} features a contact module made of a soft hemispherical elastomer with a reflective layer, with tactile images captured by an internal fisheye lens. The Insight sensor~\cite{sun2022soft} uses a thumb-like 3D conical shape and internal illumination provided by a ring of RGB LEDs that allows tactile information from multiple directions. Another variant, the 9DTact sensor~\cite{lin20239dtact} incorporates a translucent gel layer to reflect light and a black gel layer to absorb internal and ambient light. It constructs depth maps based on white light intensity, but is unable to capture the texture details of those objects. The C-sight~\cite{fan2024design} sensor enhances the 9DTact production method by using a multi-material 3D printer to fabricate the reflective and black surface layers, providing a highly repeatable assembly.


\subsubsection{\acrlong{tlb}}
The TLB method combines tactile and vision modalities to provide more comprehensive and precise tactile data (see Fig.~\ref{fig:camera_view}). Compared to the categories above, its most obvious feature is that the skin is made of transparent or translucent materials, allowing the camera to capture external light and obtain visual information such as the color or texture of the object (see Fig.~\ref{fig:optical_sensors_category}-D).

In sensors of the \gls{tlb} type, the TIRgel sensor~\cite{zhang2023tirgel} serves as an example because of its ability to transition between visual and tactile modes when changing the camera's focus. During the tactile mode, the lens-shaped elastomer leverages total internal reflection (TIR) to highlight tactile features, enabling it to capture details at a fine scale. In the visual mode, the device operates comparably to vision-in-hand technology, capturing details such as the object's color upon contact. ViTac~\cite{fan2024vitactip} also employs a transparent outer layer, allowing external light to be directly captured by the camera, thus providing it with partial visual sensing capabilities. Upon contact, the deformation of the transparent gel at the contact area creates refraction, which accentuates the contact information. 

However, the \acrlong{tlb} sensor has a drawback in that both visual and tactile data are present in the tactile images, creating interference between these two modalities. Moreover, tactile images are strongly affected by ambient light, whereas the design of most \gls{vbts} tactile sensors seeks to eliminate this to prevent noise unrelated to tactile information. Consequently, researchers have developed tactile sensors that combine the benefits of \gls{tlb} with other mechanisms (see Fig.~\ref{fig:optical_sensors_category}-B+D), which will be covered in the following.

\section{Combined classification scheme of \acrlong{vbts}s}\color{black} 


There is an increasing tendency for \glspl{vbts} to integrate multiple mechanisms to leverage the distinct advantages of each. These combinations can enhance tactile sensitivity to result in improved overall sensing or to enable new applications. Combining mechanisms can also allow for more versatile tactile sensing, by capturing a broader range of tactile information. Frequent pairings are described below.

\subsubsection{\acrlong{smb} + \acrlong{tlb}}
These \gls{smb}+\gls{tlb} tactile sensors build upon the \gls{tlb} design by incorporating markers within the transparent skin. The tactile sensor can then enhance its ability to predict both static and dynamic contact forces, while also capturing visual and proximal information through the transparent skin surrounding the markers (see Fig.~\ref{fig:optical_sensors_category}-A+D).

The \gls{smb}+\gls{tlb} mechanism is typified by the FingerVision tactile sensor~\cite{Yamaguchi19IJHR_FingerVision}, which incorporates black-dot markers on a flexible transparent surface. It monitors the displacement of these markers to correlate their shifts with applied contact forces. This design enables the sensor to capture tactile and visual information while maintaining a soft and flexible contact interface. The issue of mixing tactile and visual information was addressed by segmenting the sensing area with a transparent central circular region and an opaque surrounding area for tactile sensing~\cite{li2018camera}. More recently, SpecTac~\cite{wang2022spectac} employs UV LED lights that can be actively controlled alongside ultraviolet fluorescent markers to alternate between visual and tactile modes. This technology realizes the separation of marked tactile images and visual images at the hardware level through controllable UV light. Alternatively, MagicTac~\cite{fan2024magictac} utilizes a soft support material as both markers and filler: when external light enters the skin, it undergoes refraction, with contact areas exhibiting refraction patterns distinct from other regions. This enables MagicTac to capture partial visual information while also enhancing tactile feedback. 

\subsubsection{\acrlong{mmb} + \acrlong{tlb}}
Similarly to \gls{smb}+\gls{tlb} sensors, these \gls{mmb}+\gls{tlb} tactile sensors include morphological marker-based tactile structures within the transparent skin instead of simple markers. These structures can include biomimetic features to enhance the tactile sensor’s capabilities (see Fig.~\ref{fig:optical_sensors_category}-B+D).


FingerVision with Whiskers~\cite{yamaguchi2021fingervision} uses whiskers both as a medium for contact with objects in the environment and as markers, with each whisker extending throughout the skin layer. When external forces disturb the system, the whiskers amplify the displacement at their tips through a lever effect, thereby enhancing the sensor's sensitivity to external forces. Another example, ViTacTip~\cite{fan2024vitactip}, modifies the TacTip \gls{mmb} tactile sensor technology with a soft translucent cover filled with a soft gel~\cite{fan2024vitactip}. 

In \glspl{vbts} that integrate \gls{smb} or \gls{mmb} and \gls{tlb}, a primary difficulty is separating tactile and visual data from a single image that has combined modalities. Interference between tactile and visual data, particularly when using deep learning techniques to predict contact forces, can result in critical errors. Moreover, markers can hinder observation of the object's texture. To address these issue, work on the ViTacTip~\cite{fan2024vitactip} has introduced Generative Adversarial Networks (GANs) to manage the transition between tactile and visual modes. This technique reduced the influence of ambient lighting and object appearance on tactile information and the issue of markers that block visual data, allowing simultaneous prediction of contact forces and object texture.

\subsubsection{\acrlong{smb} + \acrlong{rlb}}
These \gls{smb}+\gls{rlb} tactile sensors use simple markers within the reflective layers. Upon contact, the camera simultaneously captures changes in light intensity and the displacement of these markers. This design preserves the sensor's ability to detect texture features while allowing a capability to predict shear displacements and forces (see Fig.~\ref{fig:optical_sensors_category}-A+C). \textcolor{black}{While \gls{smb}+\gls{rlb} sensors retain the advantages inherent to \gls{smb} and \gls{rlb} sensors, they can also introduce limitations such as reduced texture sensitivity resulting from the obscuration of some detail due to the markers.}


Several \gls{smb}+\gls{rlb} sensors embed markers within their reflective layers to enable prediction of contact shear force. The GelStereo sensor~\cite{cui2021hand}, for example, uses a stereo camera along with a stereo vision system to achieve absolute depth information, bypassing the need for tricolor structured light technology. GelSlim 3.0~\cite{taylor2022gelslim} builds on the original GelSlim design by introducing an array of black markers embedded within the outer layer. This enhancement improves its ability to track dynamic contacts and the accuracy of force prediction. The F-TOUCH sensor~\cite{li2020f, li2020f2} captures high spatial resolution images, similar to GelSight, while incorporating three black markers within its structure to predict contact forces and their vector components. In contrast, the L$^{3}$F-TOUCH sensor~\cite{li20233} employs AR tags placed at the bottom of the sensor and a set of mirrors to predict the direction and magnitude of contact forces. The UVTac sensor~\cite{kim2022uvtac} actively switches UV LEDs to decouple the images of the reflective film and the marker in two different sensing modes.

\subsubsection{\acrlong{rlb} + \acrlong{tlb}} These \gls{rlb}+\gls{tlb} tactile sensors feature a transparent outer layer integrated with a reflective layer. By adjusting the internal illumination, the sensor can actively switch between the functionalities of \gls{rlb} and \gls{tlb} sensors. This dual-mode capability combines the advantages of both tactile sensor types (see Fig.~\ref{fig:optical_sensors_category}-C+D).


There are a few examples of combining \gls{rlb} + \gls{tlb} mechanisms to transition between tactile and visual modes. The first such sensor~\cite{hogan2021seeing} incorporated a half-silvered coating on its surface and dynamically modified the internal lighting conditions. When internal illumination is intense, the sensor skin became opaque because of internal light reflection, and when internal illumination is deactivated, the skin became transparent to capture visual data. The StereoTac sensor~\cite{roberge2023stereotac} employs a similar strategy but also uses stereo cameras to reconstruct the 3D shapes of objects. Meanwhile, the surface of the VisTac sensor~\cite{athar2023vistac} is covered with a translucent film whose opacity changes depending on the intensity of light. \textcolor{black}{By modifying the internal light levels, the material's transparency can be altered, enabling rapid (<200 ms) transitions between visual and tactile operational modes.} A notable feature among all these sensors is their ability to alternate between tactile and visual modes by changing internal illumination. \textcolor{black}{Another solution involves preserving both visual and tactile modes without transition, instead employing a hybrid visual-tactile approach. In PolyTouch's~\cite{zhao2025polytouch} implementation, a fisheye lens camera's field of view (FoV) is partitioned into three distinct sections: a central non-transparent reflective layer for tactile information capture, flanked by bilateral transparent optical films that directly acquire visual data.} 

\begin{table*}
\centering
\caption{\textcolor{black}{Common Methods for Interpreting Tactile Information For \glspl{vbts}}}
\label{tab:common methods}
\begin{tabular}{c}
\includegraphics[width=1\textwidth]{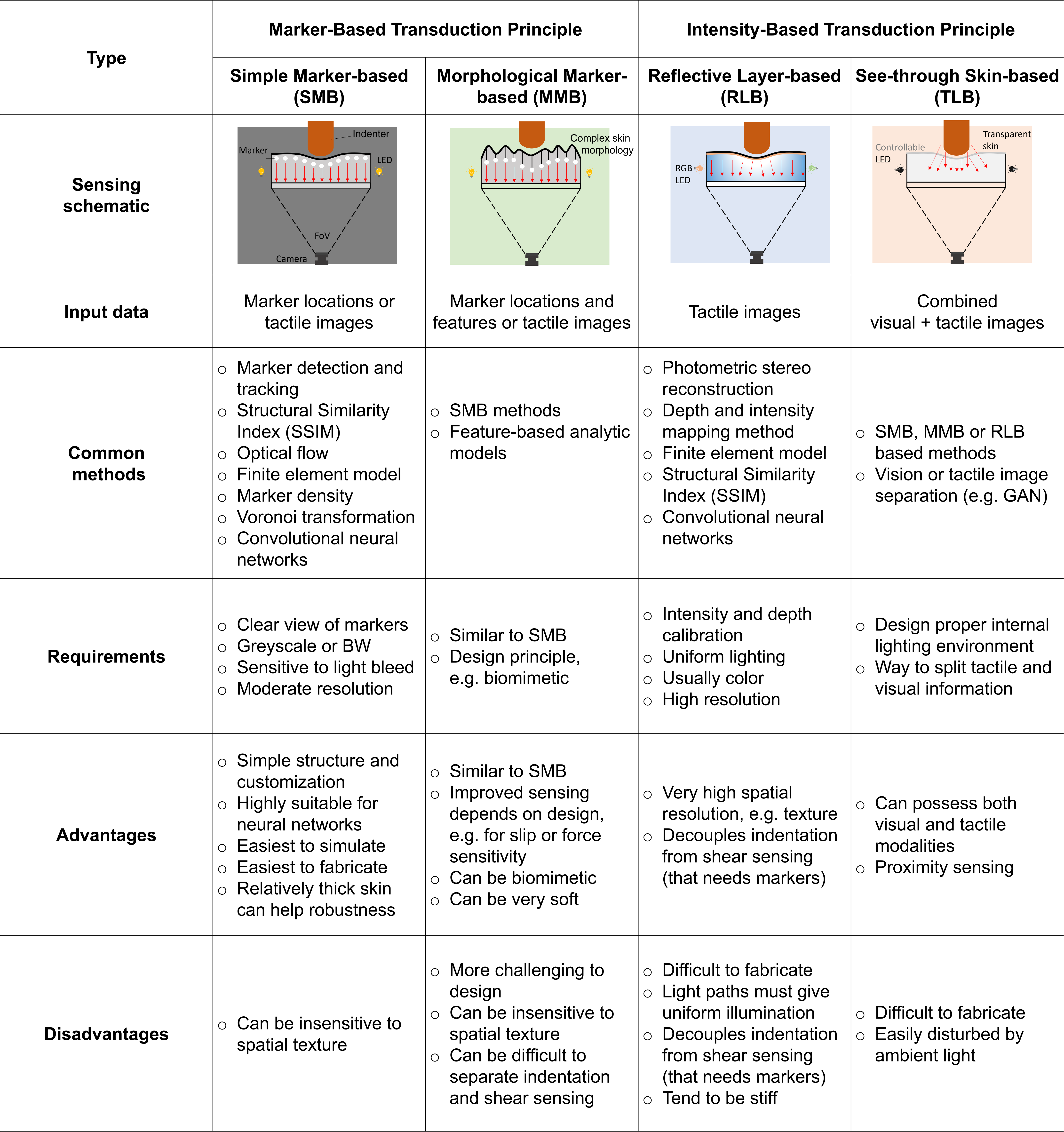} \\ 
\end{tabular}
\end{table*}

\section{Data Processing with \acrlong{vbts}s}\label{sec:modules}

A core challenge for \glspl{vbts} lies in accurately extracting contact information such as contact forces and object information from tactile image data. In this section, we discuss some interpretation methods for different types of \glspl{vbts} (see Table~\ref{tab:common methods}).

\subsubsection{Tactile Image Pre-processing}
Before analyzing \gls{vbts} images, several preprocessing steps are usually taken to enhance image quality and help analysis efficiency. These steps include cropping the image to extract the tactile sensing area, which improves both the processing speed and accuracy~\cite{li2024biotactip}. Furthermore, image down-sampling can play a crucial role in reducing computational load, which is particularly valuable for applications requiring real-time performance. In addition, image binarization~\cite{lin2023bi}, which converts grayscale images into a binary black-and-white format based on a defined threshold, is a commonly used technique to minimize noise and simplify image features, usually with \gls{smb} sensors as it helps distinguish markers from the background. Another useful pre-processing technique is image distortion calibration to correct lensing effects, for example to accurate depth and shape reconstruction of objects. Overall, these pre-processing techniques provide a foundation for advanced image analysis methods, such as deep learning, and can improve the accuracy and efficiency of data processing.

\subsubsection{Methods for Marker-Based Transduction}
The approaches to interpret tactile images for SMB and MMB sensors are separated into two categories: data-driven methods using machine learning and analytical models that incorporate computer vision technologies and physical models. We consider the benefits and challenges of these methods in the following.

\textbf{Image-based data-driven methods} are the most commonly used techniques in \glspl{vbts} for interpreting and analyzing tactile image data. \textcolor{black}{These methods often involve using neural network architectures such as Convolutional Neural Networks (CNNs)~\cite{lepora2022digitac, liu2017recent} for tasks such as object classification~\cite{zhang2025design, fan2024vitactip,fan2024magictac}, 3D shape reconstruction~\cite{10517361}, contact force prediction~\cite{xu2024vision, chen2024transforce}, object pose estimation~\cite{xu2024vision,lepora2020optimal,lloyd2023pose,lepora2021pose}, slip detection~\cite{james2018slip,james2020slip}, and contact pose prediction for tactile servoing~\cite{lloyd2023pose,lepora2021pose}.} One significant benefit of deep learning is its ability to autonomously and effectively extract relevant features from tactile images, allowing a precise prediction of contact-related physical information. 

However, image-based data-driven approaches also have disadvantages. Firstly, they require considerable data for training models, which is expensive in time and effort to gather those data. Secondly, these models can have narrow generalizability, in that a model trained on one \gls{vbts} might not be suitable for other sensors of the same type with slight differences in fabrication; also, a model developed for one task may not generalize to different tasks involving different contact data. Moreover, deep learning models are often referred to as a ``Uninterpretable model'' because they do not allow researchers to easily fine-tune and understand the models via parameter adjustents, which hinders their transparency and interoperability. Lastly, data-driven methods require significant computational resources, often relying on high-performance GPUs to train models and handle image processing tasks, which is not ideal for embodied artificial intelligence systems. Therefore, when adopting the data-driven method, one should weigh the benefits against these costs, including the investments in time, data, and computational resources.

\textbf{Marker displacement-based data-driven methods} use marker displacements as inputs for the tactile model rather than the entire tactile image. This approach discards some tactile image information, such as light intensity variations or marker shape, and focuses more on utilizing marker positions to predict contact information. Examples include using a support vector machine (SVM) on a vector of marker velocities to accurately predict slip or non-slip using a TacTip, both with a single sensor~\cite{james2018slip} and integrated with a three-finger robotic hand~\cite{james2020slip}.  Another method is to use the marker displacements as inputs to a Graph Neural Network (GNN)~\cite{fan2023tac,zhang2025design}, which has the benefit of significantly improved  computational efficiency compared to CNNs on tactile images. The drawback is that some important information relating to the contact may be lost, for example marker eccentricity can signal contact geometry information for so-called `2.5D methods'~\cite{li2023marker}. 


Compared with image-based data-driven methods, marker displacement-based data-driven methods discard potentially irrelevant information in the tactile image and only require marker position data, reducing computational complexity. Additionally, by focusing on marker displacement, sensors using these methods can be made to be highly sensitive to specific tasks, such as slip detection, with improved computational efficiency that makes them particularly suitable to real-time dynamic tasks where reaction speed is essential. However, questions remain about their performance in other scenarios; for example, important information may be lost from the tactile image. In addition, reliable marker detection and tracking algorithms can be challenging in practice~\cite{li2023marker}. Issues such as internal reflections, external light bleed, or markers merging under some deformations are commonplace. For these reasons, it can be simpler and more robust to adopt image-based methods for practical applications. 


\textbf{Analytic methods} use relatively simple and interpretable algorithms or equations to extract information from tactile images. For example, the marker displacement methods discussed above rely on blob (binary large object) detection, which operates by identifying regions in an image that differ significantly from surrounding areas in brightness, color or other attributes. In some scenarios, the contact location or force can then be estimated by calculating changes in marker density or tracking the flow of markers within a specific area~\cite{lepora2021soft}, such as by forming a Gaussian kernel density map from marker displacements~\cite{lloyd2023pose,lu2024dexitac}. A related technique is to use Voronoi tessellation to generate areas bounded by groups of markers, from which information such as contact shape can be estimated by monitoring changes in these unit areas~\cite{cramphorn2018voronoi,fan2023tac}. Another common method is optical flow, which estimates the movement of objects in a scene by establishing correspondences between markers in consecutive frames of images~\cite{pagnanelli2023model}, which relates to the perception of softness. These techniques provide high levels of interpretability, enabling easy parameter modification to tailor the model to desired use cases. Additionally, they generalize well across new designs of marker-based tactile sensors and easily extend to new tasks.

There are fewer analytic methods that directly use the tactile image and that are applicable to marker-based sensors. The most useful is the Structural Similarity Index Measure (SSIM), which gives a robust estimate of the similarity of two tactile images. The SSIM is based on a local match of the brightness, contrast, and structure, giving a perceptually robust measure compared to, for example, average pixel differences. It can be used to give an overall estimate of tactile deformation, using the dissimilarity (1-SSIM) between a tactile image and a reference undeformed tactile image. This tactile deformation can be thresholded to detect contact~\cite{zhang2024tacpalm}, or used in a feedback controller to maintain stable contact~\cite{ford2023tactile,lepora2021towards}. In general, this method requires only simple calibration prior to use and is applicable to many \glspl{vbts}, including both marker-based and other types of vision-based tactile sensor.

An alternative approach for \gls{mmb} sensors is to design the morphology such that desired tactile features, {\em e.g.} contact depth or location, can be easily measured from the tactile image. The BioTacTip~\cite{li2024biotactip} uses a pattern of 4 cover tips around each marker to signal depth from the coverage of each marker. Additionally, a simple physical model of the sensor allows light-intensity variations and marker displacement information to be transformed to contact forces, including normal and friction forces. ChromaTouch~\cite{scharff2022rapid} employs a dual-layer overlay of red and blue markers to localize contact positions. Under external force, the relative displacement between markers at the contact point induces chromatic variations in the contact region. The contact center is then localized by binarizing the image after applying a hue-based threshold. By designing and combining markers with distinct morphologies, the system effectively reduces computational overhead, eliminating the need for methods such as marker tracking or detection.


\subsubsection{Methods for \acrlong{rlb} Sensors} Similarly, methods for interpreting tactile images from \gls{rlb} tactile sensors can also be categorized into analytic and data-driven methods. In the following, we discuss representative techniques from each category.

\textbf{Analytic methods:} Traditionally, GelSight-type \gls{rlb} sensors relied on analytic photometric stereo algorithms to reconstruct the shape of the sensing surface~\cite{johnson2009retrographic}. The photometric method can accurately reconstruct the fine textural detail of the contacted surface. By capturing multiple shadings of the surface indentation from distinct known lighting directions, it is possible to solve for the surface normals and reconstruct the surface's shape. There has been a great deal of work using photometric stereo with the GelSight sensor, which we refer to dedicated reviews for more details~\cite{yuan2017gelsight,abad2020visuotactile}. 

The 9DTact sensor also employs a similar approach, but compared to GelSight technology, it uses changes in grayscale pixel intensity (rather than color intensity) to calculate contact depth, thereby reconstructing the object's shape~\cite{lin20239dtact, lin2023dtact}.  

In addition, the Structural Similarity Index Measure (SSIM) can also be applied to \gls{rlb} sensors by quantifying the similarity between the current image and the initial reference image, which indirectly reflects the magnitude of the contact force to some extent. However, this approach discards substantial contact-related information and can only provide a single-dimensional output.


\textbf{Data-driven methods:} There has been a huge amount of work using \textcolor{black}{convolutional neural networks (CNNs)} to build predictive models of contact from \gls{rlb} sensors, with early works including softness prediction and grasp success prediction from GelSight tactile images~\cite{yuan2017shape,pmlr-v78-calandra17a}. In general, there are many tasks, including contact force prediction~\cite{sun2022soft, lepora2020optimal, lin20239dtact, chen2024transforce, sferrazza2019design,yuan2017gelsight}, slip prediction~\cite{james2020slip}, shape estimation~\cite{lin20239dtact, ou2024marker, do2022densetact, luo2015novel}, texture recognition~\cite{luo2018vitac, lambeta2020digit, lambeta2024digitizing} and vision-tactile fusion~\cite{zhang2025design,fan2024vitactip}, that are well suited to such methods. \textcolor{black}{Multimodal tactile perception models employing CNNs to simultaneously identifying diverse contact information such as pose estimation, contact localization, contact force prediction, and object classification have also been proposed~\cite{xu2024vision}.} Related techniques have been used with other \gls{rlb} sensors, for example, the Insight sensor~\cite{sun2022soft} utilizes a ResNet architecture to map images onto the spatial distribution of three-dimensional contact forces on its conical tactile surface. \textcolor{black}{Recent work on using Transformer architectures enables tactile models to focus on distinct segments of the input sequence, capturing long-term temporal dependencies while simultaneously processing spatiotemporal features of tactile images~\cite{han2024learning}.} As with all data-driven methods, the challenge is to obtain the large amount of data needed for training and validation, which can require custom platforms for dataset collection and large amounts of time and effort resourcing.


\subsubsection{Methods for TLB}
For TLB type sensors, the core challenge lies in segregating tactile and visual information from the acquired images. TLB sensors can have transparent or semitransparent skins, allowing external ambient light to penetrate the contact module, so that images contain both tactile and visual information. This design increases the complexity of information processing, but also provides richer data. Most TLB sensors address this issue by actively controlling internal lighting, allowing the system to acquire tactile and visual images separately, such as STS~\cite{hogan2021seeing} and StereoTac~\cite{roberge2023stereotac}. Another approach is used by TIRgel sensors~\cite{zhang2023tirgel}, which achieve the switch between tactile and visual modalities by adjusting the focal length of the camera. In this case, the processing of the tactile or visual images proceeds as in the other methods described above.

The separation of tactile and visual information can also be achieved purely by data processing. \textcolor{black}{The image-to-image translation approach using generative adversarial networks (GANs)~\cite{zhang2025design} facilitates modality conversion between tactile and visual perception. Through reconstructing marker displacement images that more accurately reflect dynamic contact information ({\em e.g.}, contact forces) from input images containing visual and displacement markers, this technique effectively mitigates interference caused by ambient illumination and object visual features on tactile data. Furthermore, it resolves the issue of marker-induced visual occlusion, allowing sensors to acquire richer visual information} \textcolor{black}{More recently, an alternative `PolyTouch' approach has emerged that employs diffusion-based tactile models to simultaneously process multimodal tactile signals for generating action predictions~\cite{zhao2025polytouch}. Experimental results on tactile manipulation tasks have demonstrated that multimodal tactile perception-trained strategies can outperform current visuomotor policies.}





\section{Research Challenges}
Although there has been much progress in the development and use of \gls{vbts}s, many challenges remain, including those associated with their integration into robotic control systems. We now consider future research directions that may resolve existing constraints and improve the effectiveness of \gls{vbts}s.

\subsection{Challenges with \gls{vbts} Hardware}
\subsubsection{Complex Manufacturing Process}
Current mainstream \glspl{vbts} are predominantly fabricated through labor-intensive processes, including mold design and manufacturing for the elastomer, casting, demolding, marker integration, lens assembly, lighting design, and integration of the PCB board with the camera module. Among these steps, the manufacturing of the contact module is particularly complex. For example, the fabrication of the elastomer in the DIGIT version of the GelSight sensor involves multiple complex stages, from designing molds and preparing the silicone solution to vacuum degassing, casting, and demolding. Some stages are time consuming to complete manually, such as applying the reflective layer and final assembly. These manual fabrication processes greatly increase labor and time costs. Furthermore, the inevitable human errors involved in these steps lead to variations between tactile sensors even within the same fabrication batch. Such variations cause the sensors to respond differently to contact through their different mechanical properties. These inconsistencies introduce variability in tactile image data, which can lead to poor generalization from sensor to sensor for data-driven models trained on such sensor-specific data.
 
\subsubsection{Durability and Robustness}
Although researchers consistently strive to create thin and highly durable protective materials, it remains difficult to achieve a balance between longevity and sensor responsiveness. Silicone's tendency to harden over time worsens the problem, resulting in more frequent replacements of contact surfaces. Therefore, silicone durability continues to be an obstacle for some applications of \gls{vbts} technology.


{\color{black}\subsubsection{Comparison of the best designs}
It is very difficult to have a fair and comprehensive quantitative comparison between existing VBTSs due to several challenges. 1) Separation of camera and contact properties: It is difficult to separate some performance characteristics between the camera module, which can be easily interchanged, and the contact module~\cite{lepora2022digitac}. Some obvious examples include response time and spatial resolution of the tactile image, but this could extend to more subtle aspects such as slip detection. 2) Lack of standardized benchmarks: Key performance metrics such as response time, spatial resolution, and durability are often measured using different experimental setups, conditions, and definitions in the literature. 3) Diversity in design and application: The vast heterogeneity in VBTS design, including differences in sensing principles, imaging methods, mechanical structure, and target tasks, means that metrics can be highly context dependent and may not reflect generalizable trade-offs. }

\subsubsection{Integration with Robots}
Compared to other types of sensors, such as thin-film piezoresistive sensors, the hardware structure of \gls{vbts} can be bulky. This larger size presents significant challenges when integrating \glspl{vbts} into robotic systems. For example, several types of \gls{vbts} have been integrated into the fingertips of the Allegro multifingered hand~\cite{yang2024anyrotate,qi2023general}. However, the size of those tactile fingertips remain significantly larger than that of a human fingertip. However, as cameras reduce in size, smaller \gls{vbts} fingertips are becoming possible that give a more seamless integration with anthropomorphic robot hands~\cite{ford2023tactile}. Even so, the requirement for an internal camera poses further challenges~\cite{saudabayev2015sensors}, as the integrated tactile fingertip leaves little room for additional driving mechanisms, such as linkage or tendon-driven structures, limiting the potential to integrate \glspl{vbts} with other actuation methods in robotic hands. These limitations highlight the difficulties in achieving a compact and versatile tactile sensing system for robotic applications.

\textcolor{black}{There are alternatives to using cameras in \gls{vbts}s, such as photodetector arrays~\cite{mao2024bioinspired,li2022skin}, which are currently limited by their low resolution (of several millimeters) but have much faster sampling rates of 1\,kHz or more. Techniques such as microstructure diffraction patterns and interference fringe analysis offer ways to optomechanically process tactile signals before photodetection~\cite{mcgovern2023development}. Other approaches include using fiber bundle arrays to transmit the tactile image to a remote camera~\cite{schneiter1984optical, di2024using} and removing the need for a bulky camera lens~\cite{xu2025thintact}. However, none of these approaches can currently produce a \gls{vbts} suitable as a large area tactile skin for robotic integration.}

{\color{black}\subsubsection{Exploration of \gls{vbts} design space} The reason for this article is the considerable research effort being invested in exploring the designs of \glspl{vbts}, motivating a new classification of sensor types to help guide that exploration. How complete is the classification scheme in Figure~\ref{fig:optical_sensors_category}? We expect that the separation of marker-based and intensity-based is fundamental. However, the concept of markers may extend to other types of discrete feature that move or change with contact, which may result in future subdivisions if that becomes a research focus. In addition, there are other mechanisms for intensity-based transduction. The focus in Figure~\ref{fig:optical_sensors_category} was on research over the last 20 years. However, in 1984, Mott, Lee, and Nicholls proposed the first \gls{vbts} using a CCD camera to capture a high-resolution (128×256 pixels) tactile image of light internally refracted from a deformable membrane~\cite{mott_experimental_1984}. Their sensor was also integrated within one finger of a gripper mounted on a robot arm~\cite{mcclelland1987giving}. This transduction mechanism gave high-resolution intensity-based tactile images of complex shapes, using a transduction mechanism based on internal refraction that has not been explored since then. Given this gap in the recent exploration of \glspl{vbts}, one might expect that there are other important design categories that have been overlooked.}

\subsection{Challenges with Information Processing}
\subsubsection{Processing frequency} Processing frequency is a key factor that influences the response speed of tactile sensors, directly affecting their ability to detect rapid tactile changes and then respond quickly in applications that require fast reactions, such as object slippage. Many traditional electronic tactile sensors ({\em e.g.} piezorestive or capacitive) can have high sampling rates and response speeds of 1\,kHz or higher. In contrast, the response speed of optical tactile sensors is generally limited by the frame rate of the cameras used (FPS) and the computational speed of the tactile models. Most optical tactile sensors operate within a frame rate range of 30 to 90 FPS. This makes them less capable of capturing high-speed dynamic changes compared to sensors that directly output electrical signals. \textcolor{black}{Cameras with higher frame rates (hundreds of FPS) or event-based cameras are technically feasible. For example, NeuroTac~\cite{ward2020neurotac} and EveTac~\cite{funk2024evetac} employ event-based cameras to replace traditional RGB cameras and utilize specially developed tactile models to track imprinted markers, allowing real-time contact force or texture prediction at high update rates. However, their size lags behind that of conventional camera modules, which presents challenges for their integration into robotic hands. }
\subsubsection{Low Generalization Ability}
We have seen in this paper that there is a huge diversity in hardware of \glspl{vbts}, such as in the design of contact modules with variations in the presence of markers, reflective layers, and other components. This diversity leads to more opportunities to explore in the field, but gives significant differences in the tactile images generated by different systems. Hence, there is a major challenge in developing a universal interpreting method for tactile images, as techniques that are effective for one type of \gls{vbts} may not be applicable to others, which has hindered the broader development and application of \gls{vbts} technology. \textcolor{black}{Recent approaches such as using a shared trunk transformer equipped with sensor-specific encoders and task-specific decoders, seeks to capture the shared latent information across diverse sensor-task pairings~\cite{zhao2024transferable,chen2025general}.} However, work is needed to obtain perceptual performance comparable to training on a single sensor, but nevertheless such interpretation methods that can generalize between different types of \gls{vbts} hold promise.
\subsubsection{Tactile data processing costs} 
\glspl{vbts}, while providing high spatial resolution and rich tactile information, can require significant computational resources to process these data. In particular for \glspl{vbts} that uses data-driven methods to predict contact information, the process demands large precollected training datasets and advanced GPUs for network training, which can be financially and energetically costly. 

\subsubsection{Sim-to-real} 
These methods enable tactile sensors to interact with virtual objects, generating rendered tactile images and corresponding contact information, such as contact force and object features. \textcolor{black}{Currently many such simulation frameworks have been proposed~({\em e.g.}~\cite{church2022tactile,lin2022tactile, shen2024simulation}). Using training data sets collected in a simulated environment, models can be trained and applied directly to real-world scenarios without the need for extensive real-world data collection and platform construction.}
However, current physics engines struggle to accurately replicate real-world interactions, especially when simulating the internal marker movement within elastomer-based tactile sensors. This limitation poses a significant challenge for virtual training of \gls{vbts}. Developing specialized simulation environments tailored for \gls{vbts} remains an open challenge in this field, requiring advancements in both physics-based rendering and soft material modeling.


\section{Conclusion}
\glspl{vbts} have advanced robotics by providing high-resolution real-time tactile feedback, enabling robotic systems to interact more effectively with complex environments. This paper has reviewed the various types of \glspl{vbts}, categorizing them into Marker-Based and Intensity-Based Transduction principles, and further divided them into subtypes based on hardware design and sensing methods. Each category exhibits unique advantages and challenges, highlighting the versatility and potential of \glspl{vbts} in various applications, from industrial automation to medical robotics.
Although \gls{vbts} technology has shown significant promise, several challenges remain. Manufacturing methods are often labor-intensive, leading to inconsistencies and reduced generalization in data-driven models. Durability, robustness, and integration with robotic systems also present key obstacles that must be addressed for a wider adoption. Furthermore, the diversity of hardware designs complicates the development of universal methods to interpret tactile images, requiring further research to enhance generalizability.
\textcolor{black}{Future research should focus on improving the manufacturing process to reduce variability, improve sensor durability, develop more compact and versatile designs for integration into robotic systems, and multimodal sensor fusion such as including vibrational or acoustic modalities.} Furthermore, addressing the challenges of interpretation of tactile images and the use of tactile data in real-time control systems will be critical to the continued advancement of \gls{vbts} technology. By overcoming these challenges, \gls{vbts}s hold a huge potential to be an essential part of the dexterity and perception capabilities of the next generation of intelligent and interactive robots, such as fully dexterous humanoids with the manual handling capabilities of humans.

\bibliographystyle{IEEEtran}
\bibliography{sensorsbib}
\vspace{-3em}
\begin{IEEEbiography}[{\includegraphics[width=1in,height=1.25in,clip,keepaspectratio]{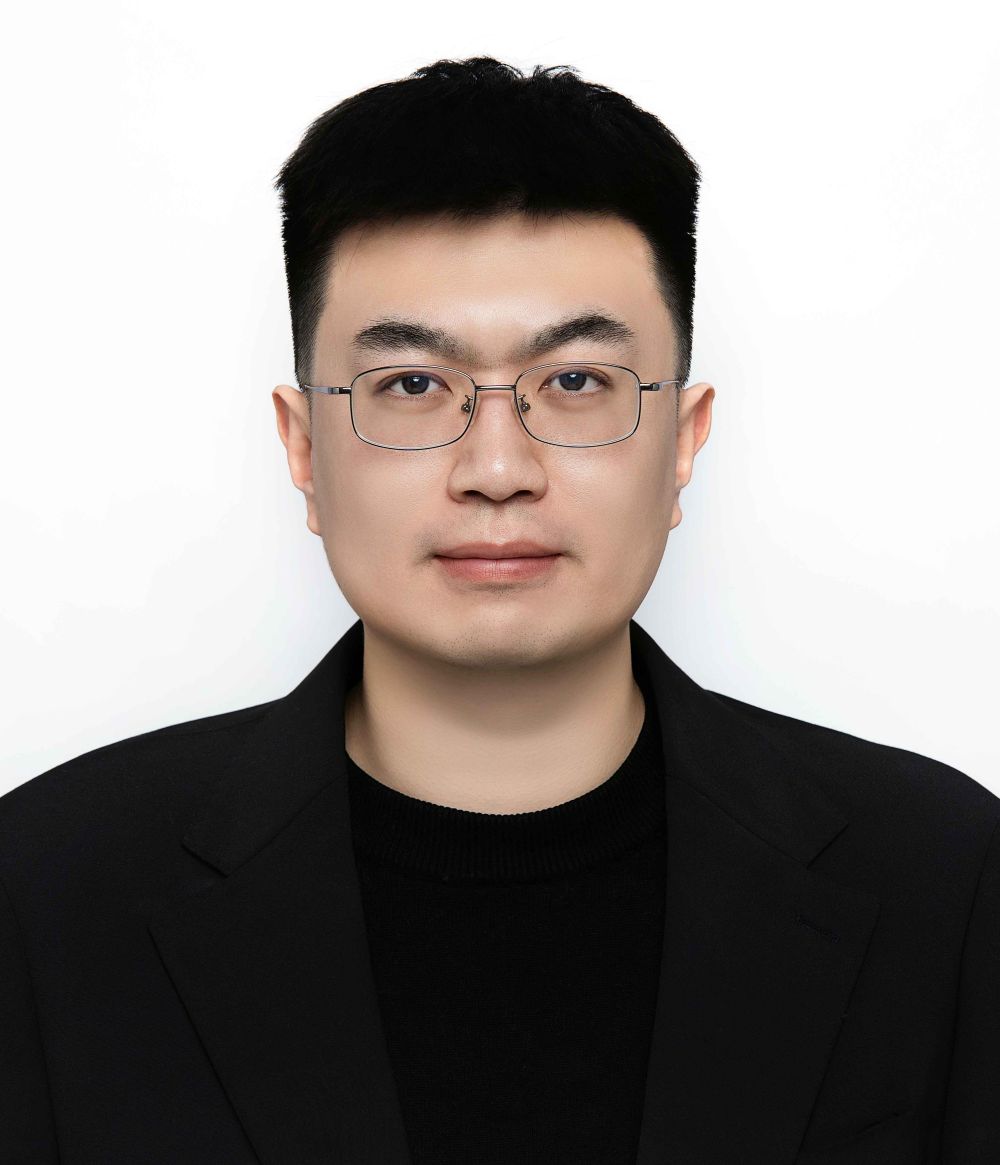}}]{Haoran Li} received his B.E. in Vehicle Engineering from Wuhan University of Technology (China) in 2019, followed by an M.Sc. in Robotics (2020) and PhD in Engineering Mathematics (2025) from the University of Bristol, UK. He is currently an Assistant Professor in the School of Robotics at Xi'an Jiaotong-Liverpool University (China). He previously served as a postdoctoral researcher at Bristol Robotics Laboratory, University of Bristol. His research focuses on tactile sensor development and electromechanical systems for dexterous robotic hands.
\end{IEEEbiography}
\vspace{-3em}
\begin{IEEEbiography}[{\includegraphics[width=1in,height=1.25in,clip,keepaspectratio]{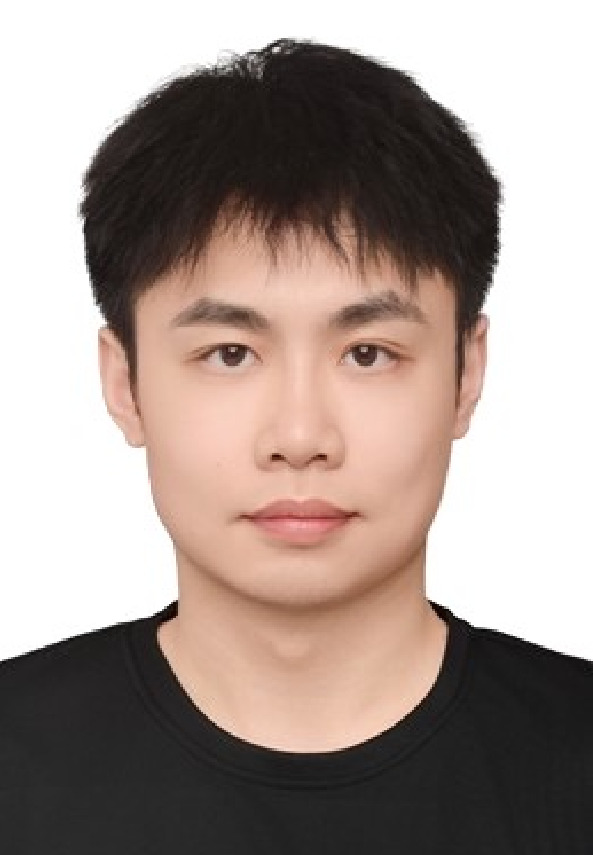}}]{Yijiong Lin} received the B.S. and M.S. degrees in Mechanical Engineering from Guangdong University of Technology, Guangzhou, China, and the Ph.D. degree in Engineering Mathematics from the University of Bristol, U.K. He is currently a Postdoctoral Research Associate with the Dexterous Robotics Group at the Bristol Robotics Laboratory, University of Bristol. His research interests include sim-to-real deep reinforcement learning, multimodal sensing and dexterous robotic manipulation. 
\end{IEEEbiography}
\vspace{-3em}
\begin{IEEEbiography}[{\includegraphics[width=1in,height=1.25in,clip,keepaspectratio]{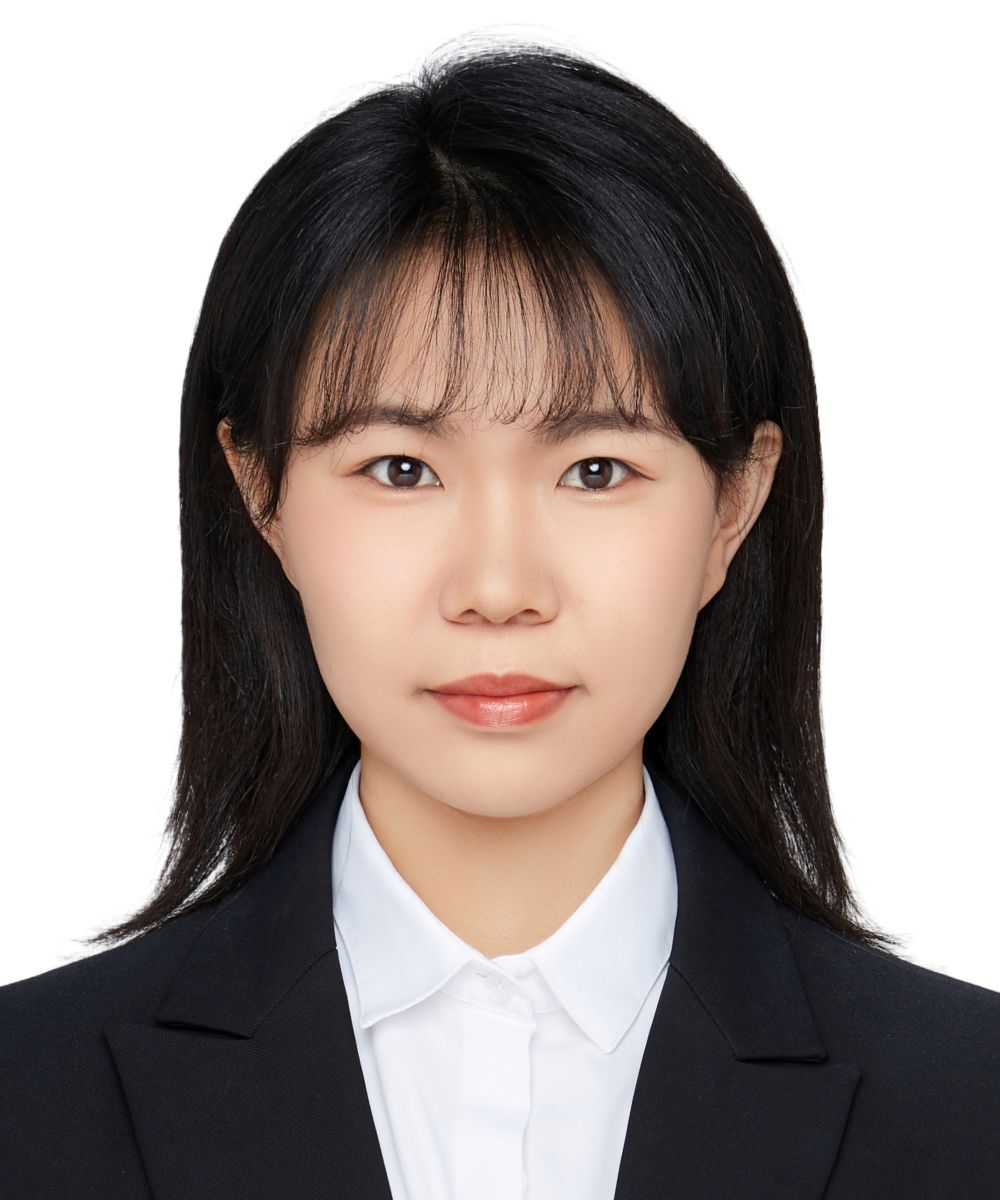}}]{Chenghua Lu} received the B.S. degree in Mechanical Engineering from Northeastern University, Shenyang, China, and the M.S. degree in Mechanical Manufacturing and Automation from the University of Chinese Academy of Sciences, Beijing, China. She is currently working toward the Ph.D. degree majoring in Engineering Mathematics with the School of Mathematics Engineering and Technology and Bristol Robotics Laboratory, University of Bristol. Her research interests include tactile sensing and soft robotics. 
\end{IEEEbiography}
\vspace{-3em}
\begin{IEEEbiography}[{\raisebox{0.3in}{\includegraphics[width=1in,clip,keepaspectratio]{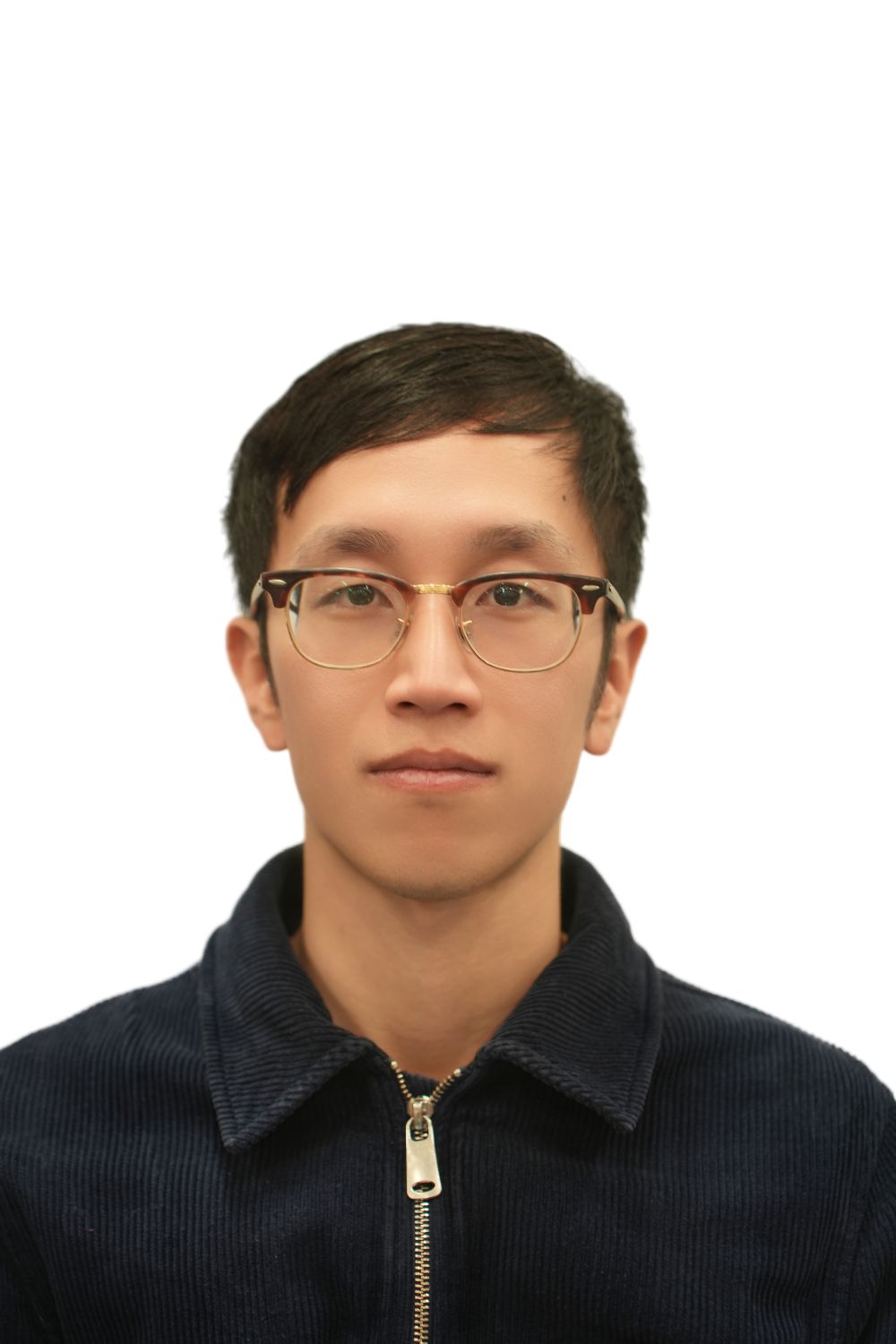}}}]{Max Yang} received his MEng degree in Aeronautical Engineering from Imperial College London, UK, in 2019. He is currently a PhD student of the Department of Engineering Mathematics and Technology at the University of Bristol and is affiliated with the Dexterous Robotics Group at Bristol Robotics Laboratory. His research interests include machine learning, dexterous manipulation, robot control, and tactile sensing.
\end{IEEEbiography}
\vspace{-3em}
\begin{IEEEbiography}[{\includegraphics[width=1in,height=1.25in,clip,keepaspectratio]{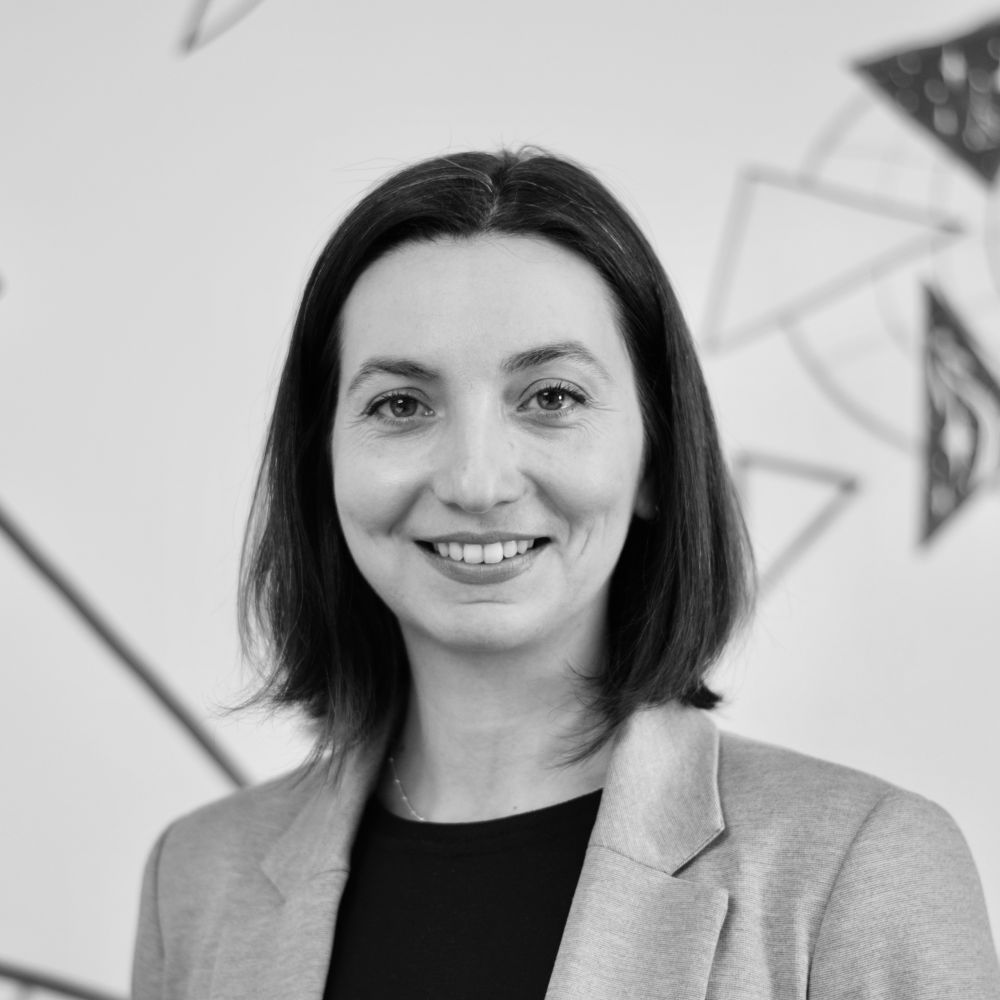}}]{Efi Psomopoulou} received the MEng degree in Electrical and Computer Engineering and the PhD degree in Robotics from Aristotle University of Thessaloniki, Greece. She is currently a Senior Lecturer at the University of Bristol and the Bristol Robotics Laboratory, U.K. She has received research funding from Horizon Europe, The Royal Society and the Advanced Research + Invention Agency. She is co-chair of the IEEE RAS Women in Engineering committee. 
\end{IEEEbiography}
\vspace{-3em}
\begin{IEEEbiography}
  [{\includegraphics[width=1in,clip, keepaspectratio]{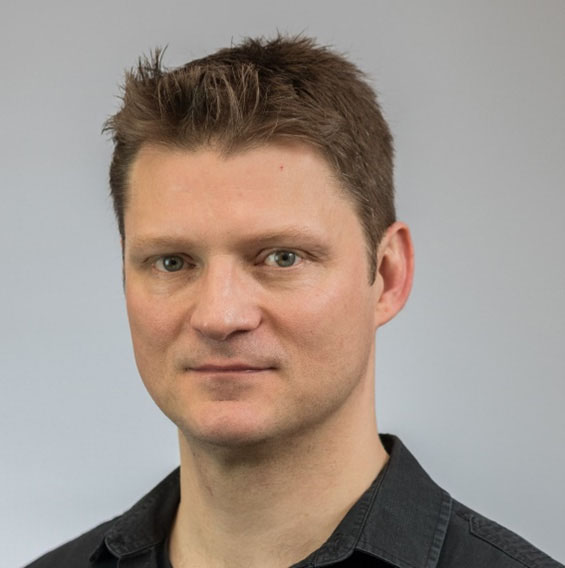}}]
  {Nathan F. Lepora} received a Ph.D. degree (1999) in Theoretical Physics from the University of Cambridge, U.K. He is currently a Professor of Robotics and AI with the University of Bristol, Bristol, U.K., leading the Dexterous Robotics Group in Bristol Robotics Laboratory, where he has been based since 2014. Prof. Lepora was a recipient of a Leverhulme Research Leadership Award on `A biomimetic forebrain for robot touch', and is a recipient of Horizon Europe and ARIA research funding on advancing robot dexterity. He has co-authored over 200 papers on robotics, neuroscience and physics, with a focus on achieving human-like dexterity by fabricating tactile robotic manipulators and applying AI. He is also interested in applying his research to understanding human intelligence. He has received an Elektra award for `Research project of the year' and a British Medical Association award for `book of the year'.
\end{IEEEbiography}

\end{document}